\documentclass[runningheads]{llncs}
\usepackage{amsmath}
\usepackage{amsfonts}
\usepackage{dsfont}
\usepackage{graphicx}
\usepackage{tikz}
\usepackage{multirow}
\usepackage{array}
\usepackage{tabularx,booktabs}
\usepackage{subcaption}
\usepackage{dirtytalk}
\usepackage{bm}
\usepackage{dsfont}
\usepackage{float}
\usetikzlibrary{arrows.meta}
\usepackage{color}
\usepackage{hyperref}
\usepackage{rotating}
\usepackage[misc]{ifsym}

\newcommand{\red}[1]{{\color{red}{#1}}}

\definecolor{cgbla}{rgb}{0.5, 0, 0.5}
\definecolor{cliver}{rgb}{0.2, 0.8, 0.85}
\definecolor{cgrasper}{rgb}{0.2, 0.85, 0.3}

\newcommand{\gbla}[1]{{\color{cgbla}{#1}}}
\newcommand{\liver}[1]{{\color{cliver}{#1}}}
\newcommand{\grasper}[1]{{\color{cgrasper}{#1}}}

\begin{document}

\title{One Patient’s Annotation is Another One's Initialization: Towards Zero-Shot Surgical Video Segmentation with Cross-Patient Initialization}

\titlerunning{One Patient’s Annotation is Another One's Initialization}

\author{Seyed Amir Mousavi$^{1,2,*}$,
Utku Ozbulak$^{1,2,*,\dagger}$(\Letter),\\
Francesca Tozzi$^{3,4}$,
Nikdokht Rashidian$^{4,5}$, 
Wouter Willaert$^{3,4}$,\\
Joris Vankerschaver$^{1,6}$, and
Wesley De Neve$^{1,2}$
}
\authorrunning{Mousavi et al.}
\institute{
$^{1}$Center for Biosystems and Biotech Data Science, Ghent University Global Campus, Incheon, Republic of Korea\\
$^{2}$IDLab, ELIS, Ghent University, Ghent, Belgium \\
$^{3}$Department of GI Surgery, Ghent University Hospital, Ghent, Belgium\\
$^{4}$Department of Human Structure and Repair, Ghent University, Ghent, Belgium\\
$^{5}$Department of HPB Surgery \& Liver Transplantation,\\ Ghent University Hospital, Ghent, Belgium\\
$^{6}$Department of Applied Mathematics, Computer Science and Statistics,\\Ghent University, Ghent, Belgium \\
(\Letter) \email{utku.ozbulak@ghent.ac.kr}
}


\maketitle

\begin{abstract}
\let\thefootnote\relax\footnotetext{* Equal contribution.}
\let\thefootnote\relax\footnotetext{${}^{\dagger}$\, Project lead.}
Video object segmentation is an emerging technology that is well-suited for real-time surgical video segmentation, offering valuable clinical assistance in the operating room by ensuring consistent frame tracking. However, its adoption is limited by the need for manual intervention to select the tracked object, making it impractical in surgical settings. In this work, we tackle this challenge with an innovative solution: using previously annotated frames from other patients as the tracking frames. We find that this unconventional approach can match or even surpass the performance of using patients’ own tracking frames, enabling more autonomous and efficient AI-assisted surgical workflows. Furthermore, we analyze the benefits and limitations of this approach, highlighting its potential to enhance segmentation accuracy while reducing the need for manual input. Our findings provide insights into key factors influencing performance, offering a foundation for future research on optimizing cross-patient frame selection for real-time surgical video analysis.
\keywords{AI-assisted surgery \and Tracking frame for segmentation \and Real-time surgical video segmentation.}
\end{abstract}

\section{Introduction}
\label{sec:introduction}

Surgical procedures are increasingly integrating artificial intelligence (AI)-based models to improve decision-making during complex minimally invasive procedures~\cite{ACS2023}. AI-driven techniques based on deep learning and computer vision are particularly valuable in real-time surgical settings, where accuracy and efficiency are crucial~\cite{Nature2022}. These models have the potential to assist clinicians in tasks such as tool tracking, anatomical structure identification, and providing real-time feedback to surgeons, reducing the risk of human errors and improving surgical outcomes~\cite{DeepLearning2022}.

A fundamental challenge in AI-assisted surgery is segmentation, which is critical for accurately identifying surgical instruments, organs, and tissues, as well as understanding their spatial relationships~\cite{Islam2020}. Traditional segmentation approaches primarily operate on individual frames, making them less effective for video-based applications due to inconsistencies in object tracking across consecutive frames~\cite{Park2020,ronneberger2015u}. Video Object Segmentation (VOS), a paradigm that ensures temporal consistency in object tracking, has the potential to transform segmentation for real-time surgeries by maintaining coherent segmentation across frames and reducing fluctuations~\cite{cheng2022xmem,ravi2024sam}. By leveraging temporal information, VOS enables more stable and reliable segmentation, addressing the limitations of frame-by-frame approaches and enhancing real-time surgical guidance~\cite{Liu2020,ravi2024sam}.

One of the major challenges in applying VOS to real-time surgical settings is the initialization of the tracked object~\cite{Wu2024}. Typically, VOS models require user input -- such as a segmentation mask, key points, or a bounding box -- to define the object that will be tracked throughout the video~\cite{Wu2024}.. However, this human-in-the-loop approach is impractical in the operating room, where maintaining strict sterilization protocols and minimizing manual intervention are critical~\cite{Hashimoto2018}. The requirement for manual initialization poses a substantial barrier to seamless AI-assisted surgical workflows, as it disrupts the autonomy and efficiency needed for real-time applications~\cite{Hashimoto2018}.

In this work, we aim to address this challenge by proposing an unconventional and innovative approach: leveraging annotations from other patients as tracking frames. Instead of requiring real-time user input, our method utilizes previously annotated surgical videos or pre-existing segmentation maps from similar cases to initialize and guide the tracking process. By transferring knowledge from prior cases, our approach aims to eliminate the need for manual intervention while maintaining segmentation accuracy and temporal consistency. Through extensive experiments on multiple surgical videos, our findings indicate that the proposed method can achieve comparable performance to using patient-specific tracking frames and, in some cases, even surpass it. Furthermore, we provide a comprehensive analysis of the benefits and limitations of our approach, highlighting avenues for future research and potential enhancements.

\section{Methodology}

\begin{figure}[t!]
\centering
\includegraphics[width=0.24\textwidth]{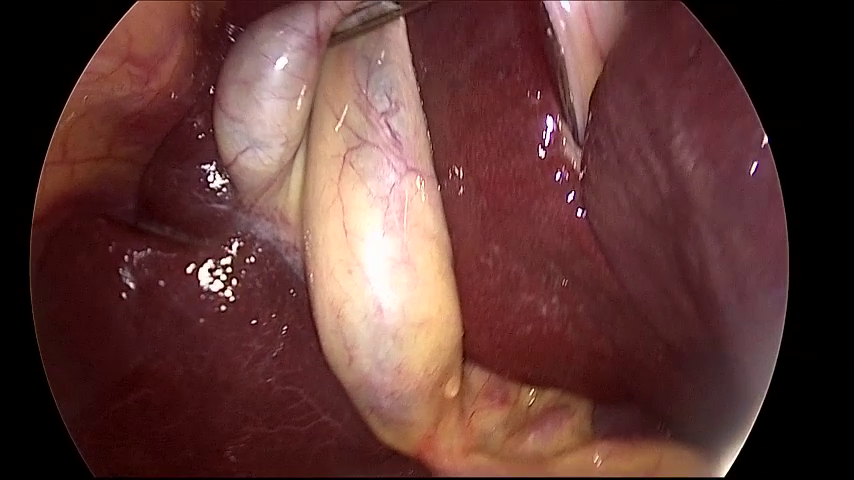}
\includegraphics[width=0.24\textwidth]{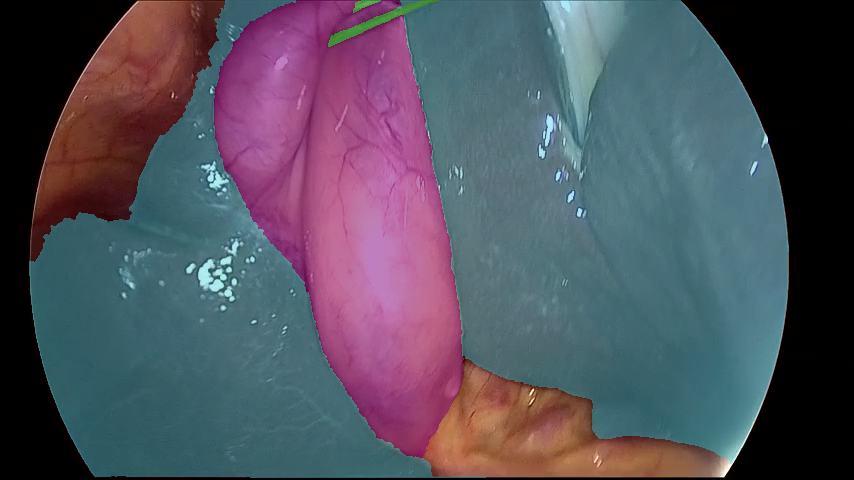}
\includegraphics[width=0.24\textwidth]{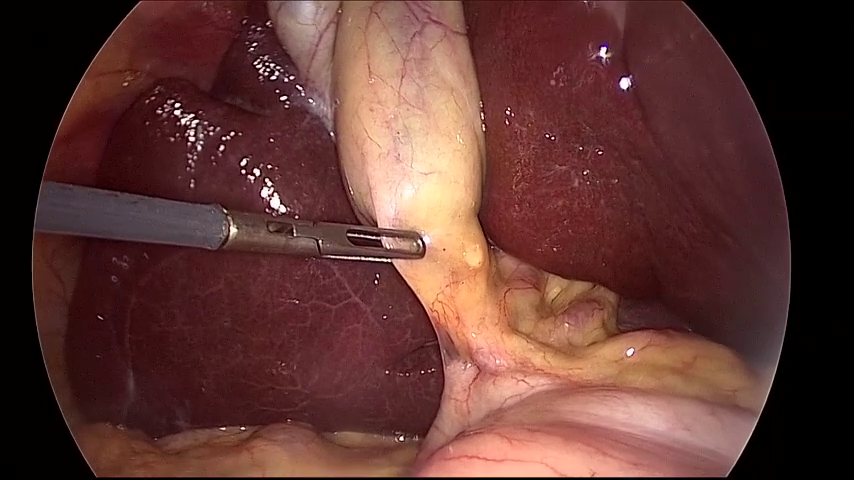}
\includegraphics[width=0.24\textwidth]{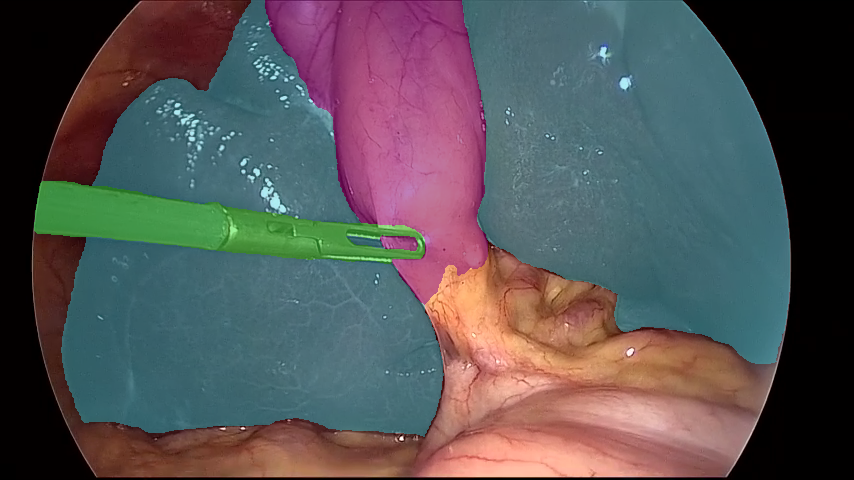}
\caption{Example images from the CholecSeg8k dataset and their annotations for the gallbladder (\textbf{\gbla{purple}}), liver (\textbf{\liver{light blue}}), and grasper (\textbf{\grasper{green}}).}
\label{fig:dataset_examples}
\end{figure}

\subsection{Model}

In this study, we utilize SAM2.1 Hiera Large~\cite{ravi2024sam}, a state-of-the-art zero-shot video object segmentation model, for real-time surgical video segmentation. Among the SAM2 variants, the Hiera Large model offers the highest segmentation accuracy that generalizes to complex scenes, including surgical videos~\cite{yu2024sam}.

Unlike conventional models that rely on task-specific fine-tuning, SAM2, trained using self-supervised learning~\cite{ozbulak2023know}, generalizes effectively to unseen surgical scenarios, accurately identifying and segmenting key anatomical structures and surgical instruments in videos. The model, built on a vision transformer architecture~\cite{vit,attention}, employs a tokenization process to identify important visual features~\cite{kang2024identifying}, enabling robust segmentation even in challenging surgical environments, achieving state-of-the-art results on several surgical video segmentation benchmarks~\cite{yu2024sam}.

The SAM2 model operates on a straightforward yet effective principle. Initially, a reference frame is selected, within which the object of interest is manually identified using (a) a segmentation mask, (b) a set of key points, or (c) a bounding box. Once this initial annotation is provided, the model propagates the object’s identity throughout the subsequent frames, maintaining temporal consistency in segmentation~\cite{ravi2024sam}. This tracking mechanism allows SAM2 to adapt to variations in object appearance, motion, and occlusions, making it well-suited for dynamic surgical environments where objects may undergo deformation, partial obstruction, or illumination changes. Among these properties, our work specifically investigates the selection and impact of the reference frame, analyzing how different choices may influence segmentation accuracy~\cite{yu2024sam}.

\begin{table}[t!]
\centering
\scriptsize
\caption{Total number of frames used in each video for segmentation analysis, categorized by target objects (gallbladder, liver, and surgical grasper).}
\setlength{\tabcolsep}{4pt}
\renewcommand{\arraystretch}{1}
\label{tbl:data}
\begin{tabular}{lcccccccccc}
\toprule
Objects      & V1 & V12 & V17 & V18 & V20 & V24 & V37 & V48 & V52 & V55 \\
\midrule
Any          & 1,280    & 640      & 320      & 160      & 160      & 960     & 480     & 240     & 800     & 240    \\
\midrule
Gallbladder  & 992     & 624      & 320      & 160      & 160      & 686     & 479     & 219     & 800     & 240    \\
Liver        & 1,280    & 640      & 320      & 160      & 160      & 960     & 480     & 240     & 800     & 240    \\
Grasper      & 849     & 577      & 240      & 160      & 160      & 400     & 480     & 240     & 800     & 240    \\
\bottomrule
\end{tabular}
\end{table}

\subsection{Data}

For our experiments, we use the CholecSeg8k dataset, which is sampled from the Cholec80 endoscopic video dataset~\cite{Czempiel2022}. The Cholec80 dataset consists of 80 laparoscopic cholecystectomy videos recorded at the University Hospital of Strasbourg. CholecSeg8k is a frame-by-frame annotated subsample of 17 videos from Cholec80, encompassing 13 categories. These include key anatomical structures such as the gallbladder and liver, secondary structures like fat and connective tissue, and surgical tools such as the grasper.

In this work, we use 10 of those 17 videos and among the 13 target categories, we specifically use three target objects: the liver, gallbladder, and the grasper, a surgical tool characterized by considerable movement compared to other structures. Detailed frame counts for these datasets are provided in Table~\ref{tbl:data} and a set of example images with their annotation overlays is provided in Figure~\ref{fig:dataset_examples}. 

\subsection{Approach}

As briefly described in Section~\ref{sec:introduction}, our work focuses on the problem of real-time surgical object segmentation. Our goal is to eliminate the human-in-the-loop requirement, enabling the segmentation model to function autonomously without manual intervention. To achieve this, we explore a novel hypothesis: using annotations from other patients as reference frames for tracking key anatomical structures and surgical tools.

To utilize SAM2 or any other VOS model, users must provide a reference frame that initializes the tracked object and enables consistent segmentation throughout the video. 

Our approach is based on a simple yet powerful and intuitive observation: in most surgical videos, key anatomical structures visually appear approximately similar across different patients. Leveraging this property, we investigate the feasibility of using a reference frame from one patient, along with its annotation, to segment similar anatomical structures in other patients. To improve the clarity of this approach, we introduce the following terminology:

\textbf{Donor patient:} The patient from whom the reference frame and corresponding segmentation mask are obtained.

\textbf{Recipient patient:} The patient whose surgical video receives the aforementioned reference frame and the segmentation mask for object tracking and segmentation.

Of the 10 selected videos from the CholecSeg8k dataset, we designate 5 as donor patients (Videos 1, 17, 20, 24, 55) and the remaining 5 as recipient patients (Videos 12, 18, 37, 48, 52). For each donor-recipient pair, we randomly sample ten reference frames from the donor patient’s video, ensuring diversity in anatomical views, lighting conditions, and surgical phases. Each selected reference frame, along with its corresponding annotation, is then used to initialize object tracking in the recipient patient’s video. This approach leads to 50 object tracking outcomes for a single recipient patient. 

\subsection{Evaluation}

We evaluate the segmentation performance for a donor-recipient pair using the average intersection-over-union (IoU) for the entire recipient video~\cite{zhou2019iou}. Given a segmentation prediction $\texttt{S}$ and the ground truth $\texttt{Y}$ with $\texttt{S}, \texttt{Y} \in \{0,1\}^{H\times W}$, IoU is defined as $\text{IoU}(\texttt{S}, \texttt{Y}) = \frac{|\texttt{S} \cap \texttt{Y}|}{|\texttt{S} \cup \texttt{Y}|}$ where the numerator represents the number of pixels correctly classified as belonging to the segmented object (true positives), and the denominator accounts for all pixels identified as part of the object in either the prediction or the ground truth (true positives + false positives + false negatives).

\begin{table}[t!]
\centering
\scriptsize
\caption{Segmentation performance of SAM2 for different anatomical structures in surgical videos. The table presents the IoU scores for the gallbladder, liver, and surgical grasper across multiple recipient videos. Performance is evaluated using a baseline IoU (benchmark with reference frames from the same patient), average IoU (mean IoU across all 50 donor frames) and its standard deviation, and best IoU(highest IoU achieved among all donors). Results using a donor frame that leads to a greater IoU than the baseline are highlighted with bold font. Empty cells represent non-existent objects in those video segments.}
\setlength{\tabcolsep}{2.2pt} 
\renewcommand{\arraystretch}{1.1} 
\label{tbl:iou_table}
\begin{tabular}{lcccc|cccc|cccc}
\toprule
\multirow{2}{*}{\shortstack{Video\\Segment}} & \multicolumn{4}{c}{Gallbladder} & \multicolumn{4}{c}{Liver} & \multicolumn{4}{c}{Grasper} \\
\cmidrule[0.5pt]{2-13}
~ \phantom{---} & Baseline & Avg & Std & Best & Baseline & Avg & Std & Best & Baseline & Avg & Std & Best \\
\midrule
V1 S1 & 96.1 & 41.9 & 30.1 & 71.5 & 89.9 & 91.0 & 13.2 & \textbf{94.5} & 74.7 & 50.7 & 24.6 & \textbf{80.4} \\
V1 S2 & 96.2 & 27.7 & 20.6 & 56.2 & 96.7 & 90.5 & 13.0 & 94.7 & 86.9 & 82.7 & 21.2 & \textbf{88.9} \\
V1 S3 & 94.7 & 46.0 & 28.2 & 70.2 & 96.1 & 58.2 & 29.4 & 94.2 & 86.9 & 49.9 & 27.6 & 72.7 \\
V1 S4 & 81.8 & 16.7 & 21.3 & 58.2 & 95.4 & 88.1 & 18.1 & 92.3 & 82.7 & 64.6 & 9.4  & 69.9 \\
V1 S5 & 86.7 & 41.4 & 24.1 & 61.4 & 91.7 & 82.0 & 27.3 & \textbf{93.9} & 80.7 & 74.5 & 1.2  & 76.9 \\
V1 S6 & --   & --  & -- & -- & 94.3 & 52.4 & 38.2 & 89.8 & --   & -- & -- & -- \\
\midrule
V17 S1 & 95.2 & 86.7 & 25.9 & \textbf{95.6} & 93.0 & 67.1 & 17.9 & 85.9 & 72.7 & 58.5 & 32.8 & \textbf{78.2} \\
V17 S2 & 95.3 & 62.7 & 42.4 & 94.7 & 93.0 & 55.8 & 25.6 & 90.9 & 87.6 & 67.2 & 34.1 & 86.0 \\
V17 S3 & 64.9 & 13.5 & 26.4 & \textbf{73.1} & 91.7 & 26.1 & 35.5 & 88.4 & --   & --   & --   & \textbf{--} \\
\midrule
V20 S1 & 92.8 & 32.6 & 28.2 & 87.0 & 96.2 & 78.6 & 26.0 & 95.4 & 94.6 & 53.4 & 46.7 & 94.3 \\
\midrule
V24 S1 & 88.7 & 7.0  & 6.6  & 15.9 & 92.4 & 61.9 & 37.4 & \textbf{92.5} & --   & --   & --   & -- \\
V24 S2 & --   & --   & --   & --  & 95.9 & 7.5  & 25.9 & 94.5 & --   & --   & --   & -- \\
V24 S3 & 94.0 & 79.2 & 32.9 & \textbf{94.2} & 95.3 & 67.0 & 34.6 & 93.8 & 55.4 & 56.2 & 17.4 & \textbf{88.5} \\
V24 S4 & 76.9 & 58.9 & 11.7 & \textbf{81.7} & 94.3 & 71.4 & 31.3 & 92.9 & 86.6 & 56.8 & 20.0 & 86.4 \\
V24 S5 & 95.9 & 35.1 & 23.2 & 52.0 & 95.7 & 82.9 & 1.6  & 83.8 & 90.6 & 85.3 & 17.7 & 89.8 \\
\midrule
V55 S1 & 88.4 & 33.2 & 38.5 & \textbf{88.6} & 74.6 & 18.4 & 11.5 & 60.7 & 83.1 & 55.5 & 36.6 & \textbf{83.6} \\
\bottomrule
\end{tabular}
\end{table}

\section{Experimental Results}

In Table~\ref{tbl:iou_table}, we present the segmentation performance of SAM2 for different target objects -- gallbladder, liver, and surgical grasper -- across multiple surgical videos of recipient patients. Descriptions for the columns of this table are as follows:

\underline{Baseline} serves as a benchmark for the baseline average IoU in segmentation performance when the reference frame is taken from the recipient patient's own video (the first frame), representing an upper-bound benchmark.

\underline{Best} represents the highest average segmentation performance achieved among all donor frames across 50 cases, highlighting the outcome of the most effective donor-recipient pairing for a target object.

\underline{Avg} represent the mean performance across all tested donor frames, providing insight into the generalizability of the approach.

\underline{Std} quantifies the variability in segmentation performance, indicating the stability of the IoU when different donor frames are used.

\textbf{Average performance}. The segmentation performance using donor frames varies across anatomical structures, with the liver achieving the highest average IoU scores, followed by the gallbladder and grasper.

\textbf{Variation in results}. The variability in segmentation performance across donor frames is evident in the standard deviation (Std) values, particularly for the surgical grasper. Higher Std values indicate that segmentation performance fluctuates substantially depending on the choice of donor frames.

\begin{figure}[t!]
\centering
\begin{subfigure}{0.32\textwidth}
\includegraphics[width=1\textwidth]{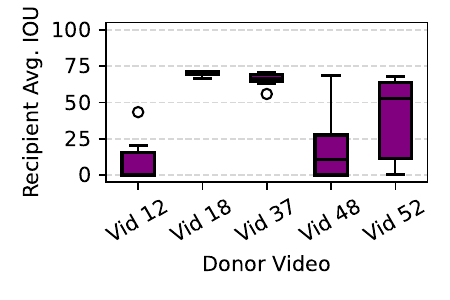}
\caption{Gallbladder}
\end{subfigure}
\begin{subfigure}{0.32\textwidth}
\includegraphics[width=1\textwidth]{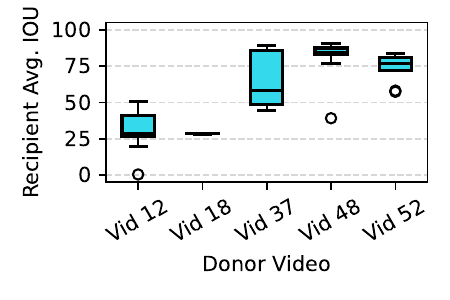}
\caption{Liver}
\end{subfigure}
\begin{subfigure}{0.32\textwidth}
\includegraphics[width=1\textwidth]{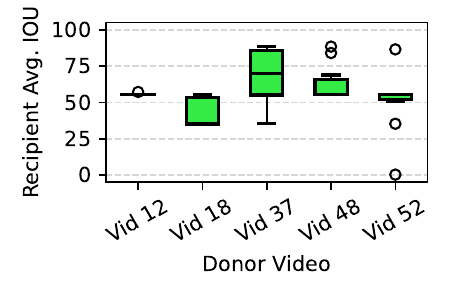}
\caption{Grasper}
\end{subfigure}
\caption{The impact of donor frame selection on the segmentation performance obtained for a recipient video, measured by average IoU, is illustrated using boxplots for (a) Gallbladder, (b) Liver, and (c) Grasper.}
\label{fig:donor_variance}
\end{figure}

\textbf{Best-performance and baseline}. Interestingly, in some cases, donor-based segmentation achieves better performance than the baseline IoU, which represents the upper-bound performance using reference frames from the same patient. We observe that 9 out of 16 recipient videos contain a case where better results are achieved for one of the target objects (5 gallbladders, 3 livers, 5 graspers) with a tracking frame from a donor video, rather than their own tracking frame, indicating that under optimal donor-recipient pairings, donor-based segmentation has the potential to exceed the expected same-patient performance.

Apart from these cases with the best result surpassing the baseline, the majority of the results for liver and grasper are close to the baseline results. However, for gallbladder, we observe a noticeable drop in the segmentation accuracy, especially for Video 1. We will discuss such cases in the upcoming section.

\textbf{Frame selection from the same donor videos}. In Figure~\ref{fig:donor_variance}, we present boxplots illustrating the impact of donor frame selection on segmentation performance across a recipient video, measured by average IoU for the gallbladder, liver, and surgical grasper. Specifically, Figure~\ref{fig:donor_variance} highlights the substantial variability in segmentation performance based on the selected donor frame, even within the same video. We can observe that, despite originating from the same patient and same surgical context, different frames yield highly inconsistent IoU scores, revealing the sensitivity of segmentation accuracy to donor tracking frame selection.

\begin{figure}[t!]
\centering
\begin{subfigure}{0.23\textwidth}
\includegraphics[width=1\textwidth]{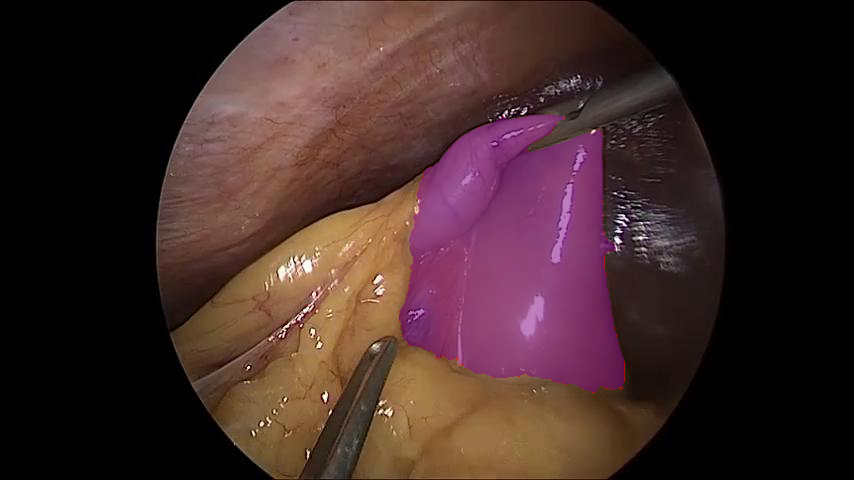}
\includegraphics[width=1\textwidth]{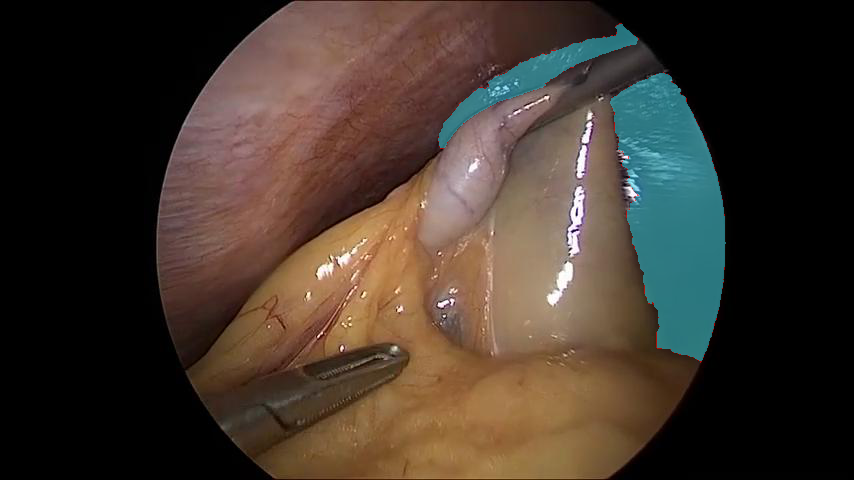}
\includegraphics[width=1\textwidth]{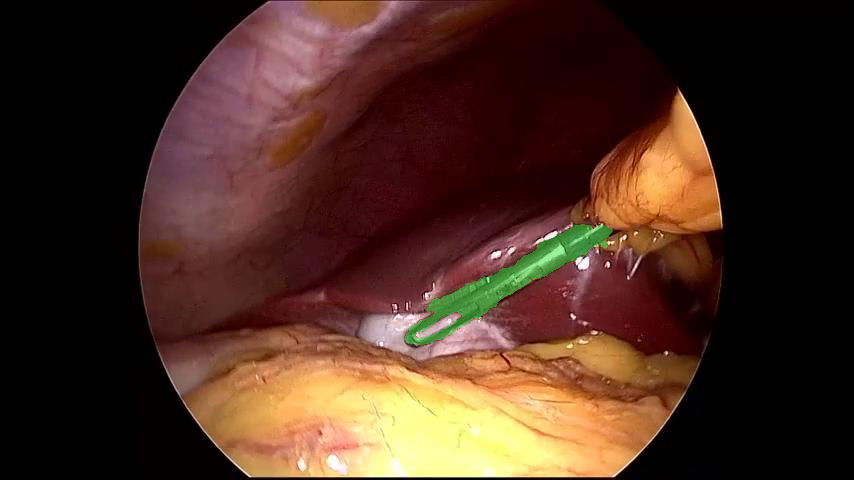}
\caption{Tracking mask}
\end{subfigure}
\begin{subfigure}{0.72\textwidth}
\includegraphics[width=0.32\textwidth]{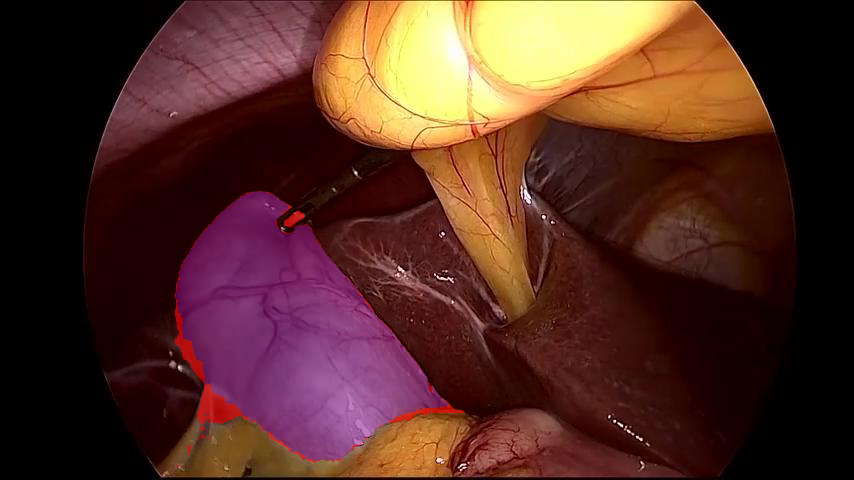}
\includegraphics[width=0.32\textwidth]{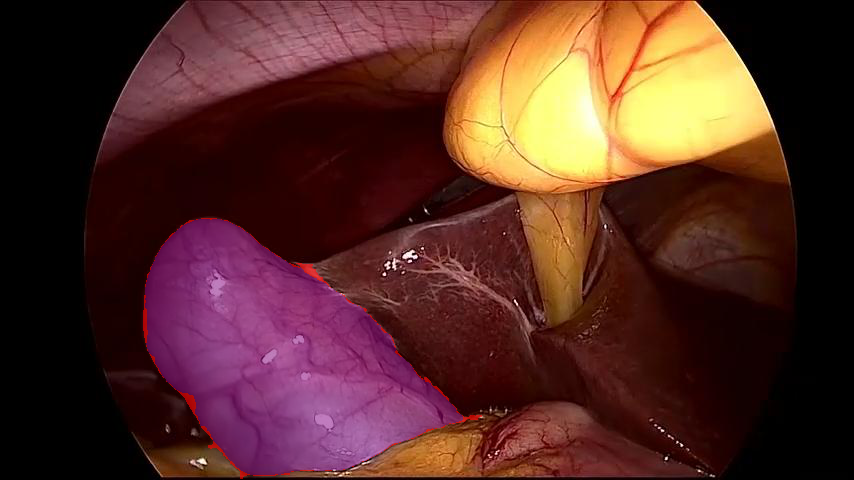}
\includegraphics[width=0.32\textwidth]{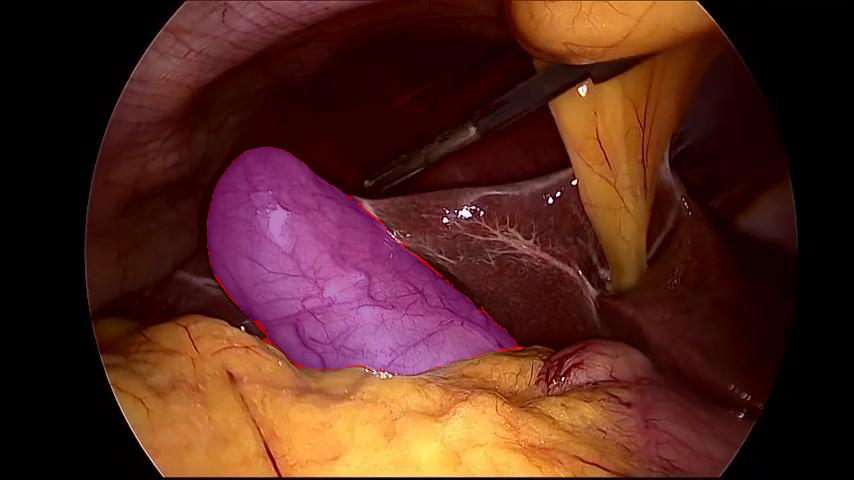}
\includegraphics[width=0.32\textwidth]{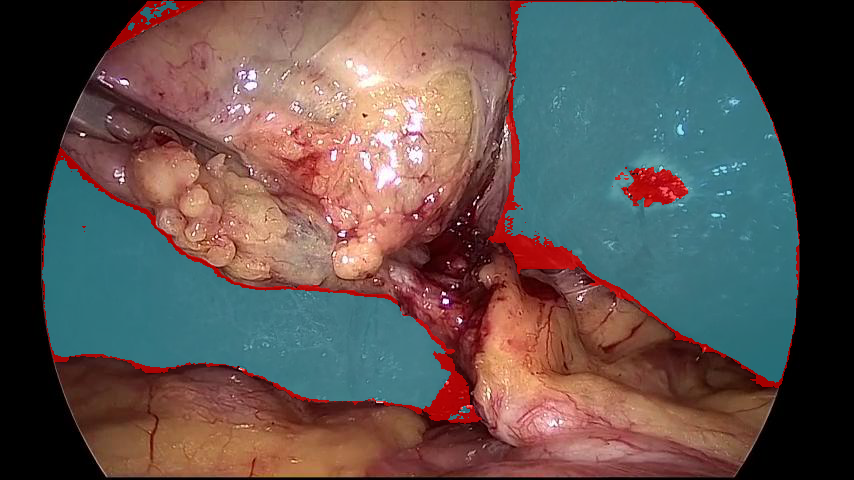}
\includegraphics[width=0.32\textwidth]{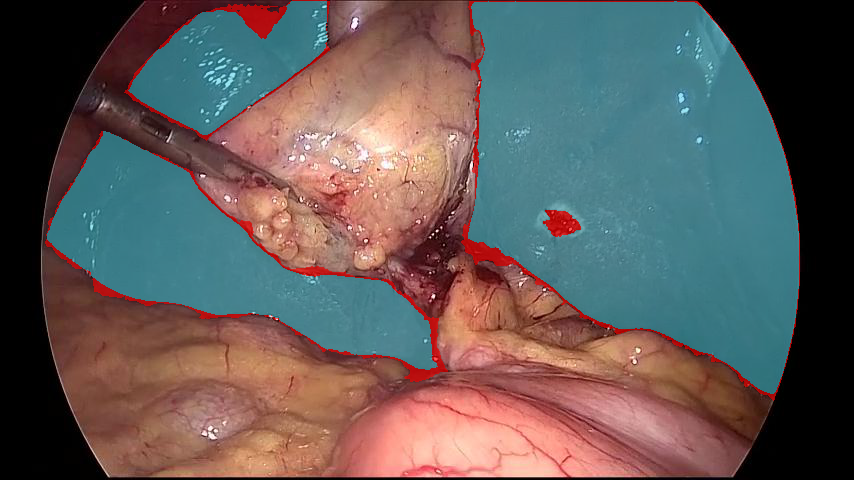}
\includegraphics[width=0.32\textwidth]{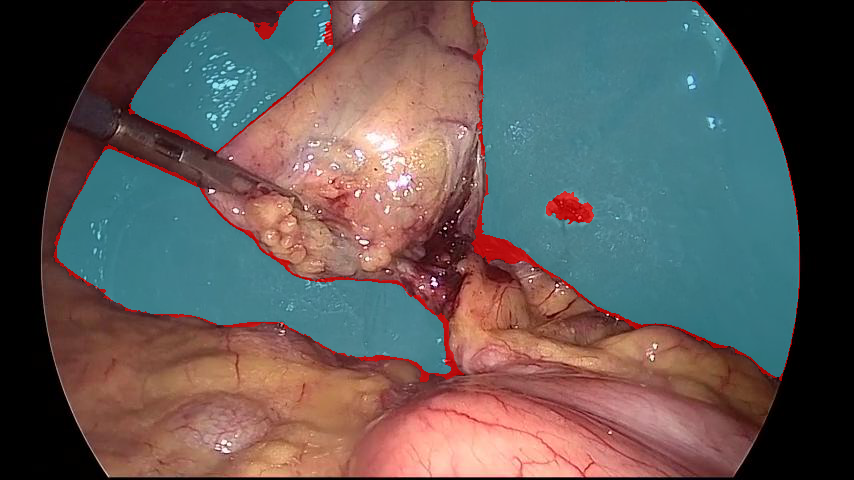}
\includegraphics[width=0.32\textwidth]{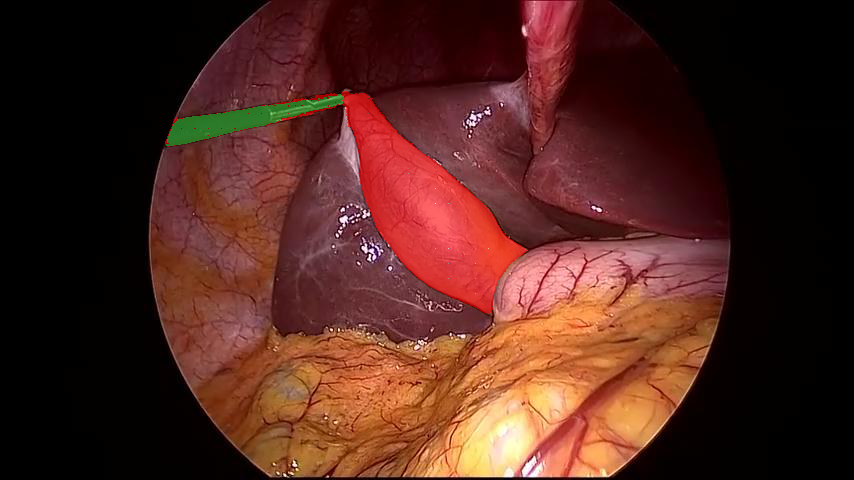}
\includegraphics[width=0.32\textwidth]{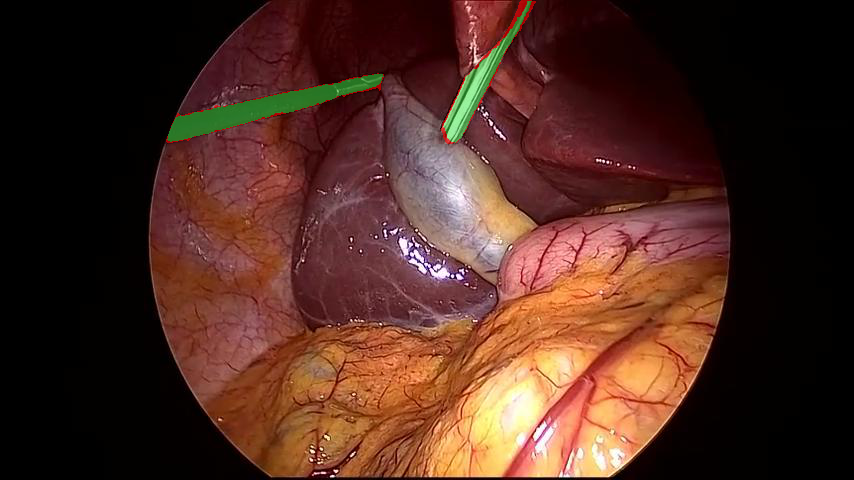}
\includegraphics[width=0.32\textwidth]{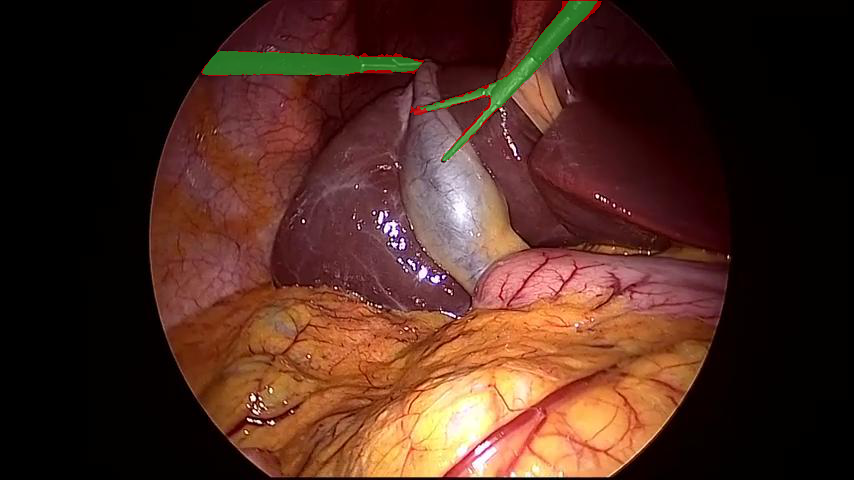}
\caption{Predictions on recipient patient videos}
\end{subfigure}
\caption{(a) Tracking masks obtained from donor patients and (b) segmentation predictions obtained from SAM2 using tracking masks for recipient patients based on the tracked objects. The first row corresponds to the \textbf{\gbla{gallbladder}}, the second row to the \textbf{\liver{liver}}, and the third row to the \textbf{\grasper{surgical grasper}}. Incorrect predictions (false negatives and false positives) are highlighted in \textbf{\red{red}}. Tracked objects in recipient patients obtain an average \textbf{IOU score $\mathbf{>0.9}$} for their respective videos, demonstrating SAM2's ability to transfer tracking information across different patients.}
\label{fig:good_examples}
\end{figure}


\section{Discussion}

\begin{figure}[t!]
\centering
\begin{subfigure}{0.23\textwidth}
\includegraphics[width=1\textwidth]{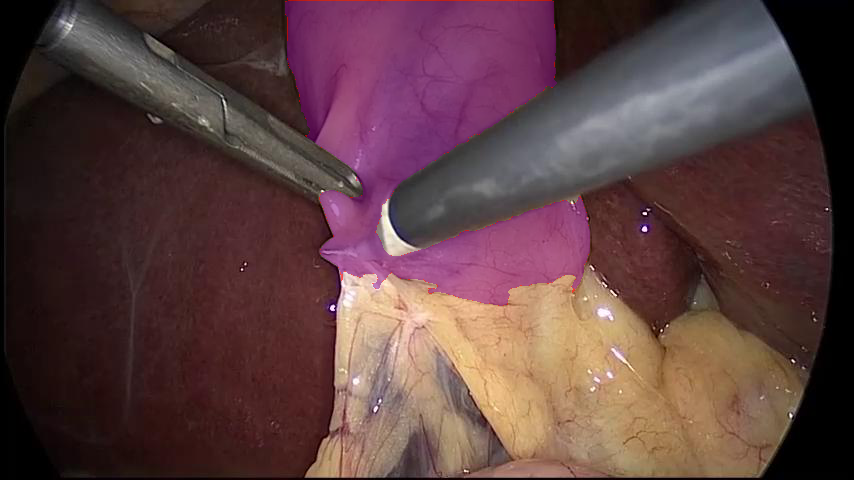}
\includegraphics[width=1\textwidth]{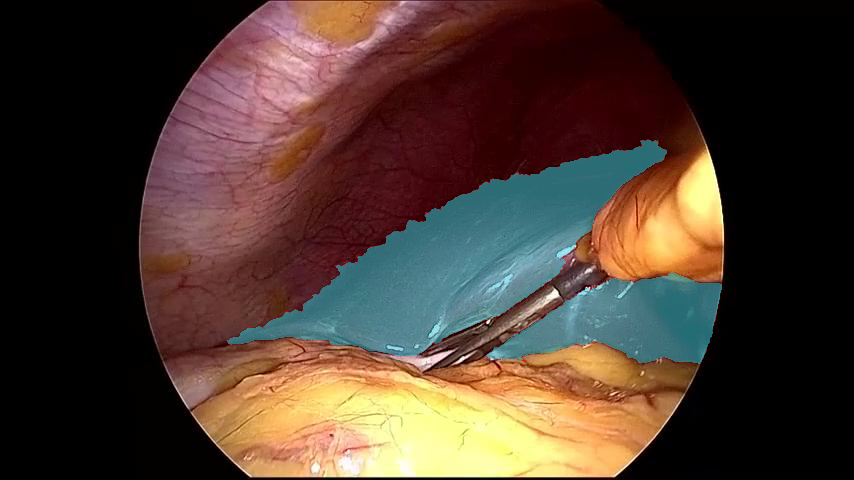}
\includegraphics[width=1\textwidth]{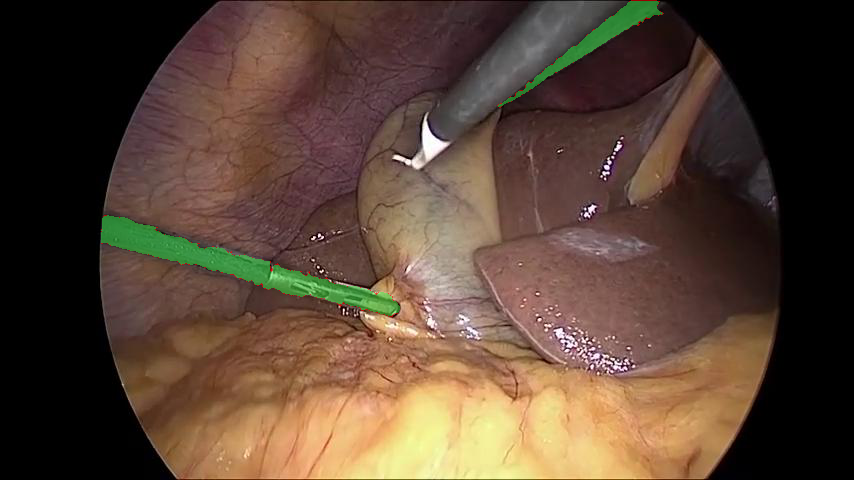}
\caption{Tracking mask}
\end{subfigure}
\begin{subfigure}{0.72\textwidth}
\includegraphics[width=0.32\textwidth]{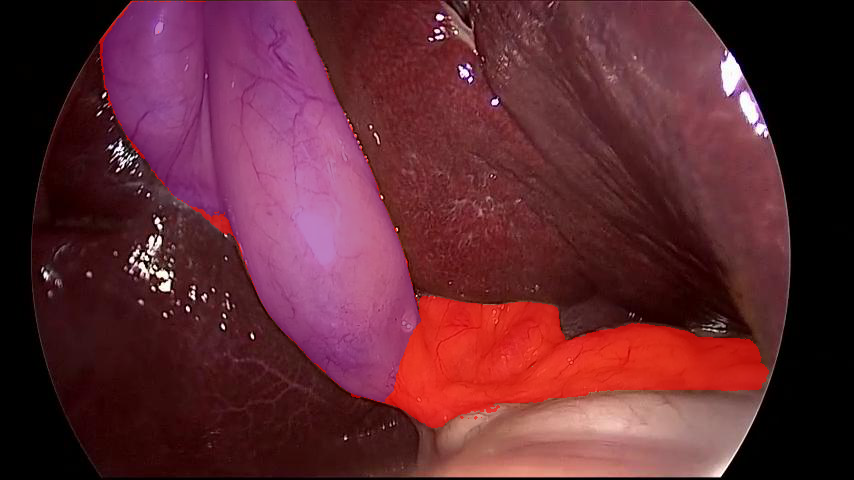}
\includegraphics[width=0.32\textwidth]{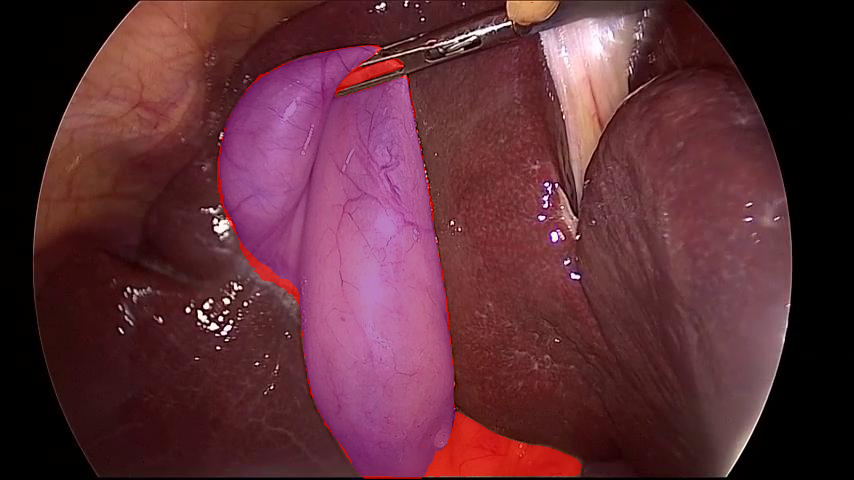}
\includegraphics[width=0.32\textwidth]{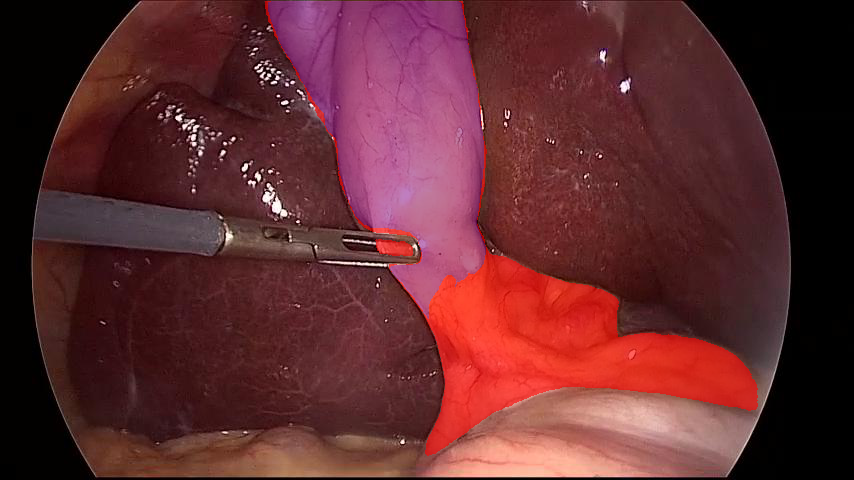}
\includegraphics[width=0.32\textwidth]{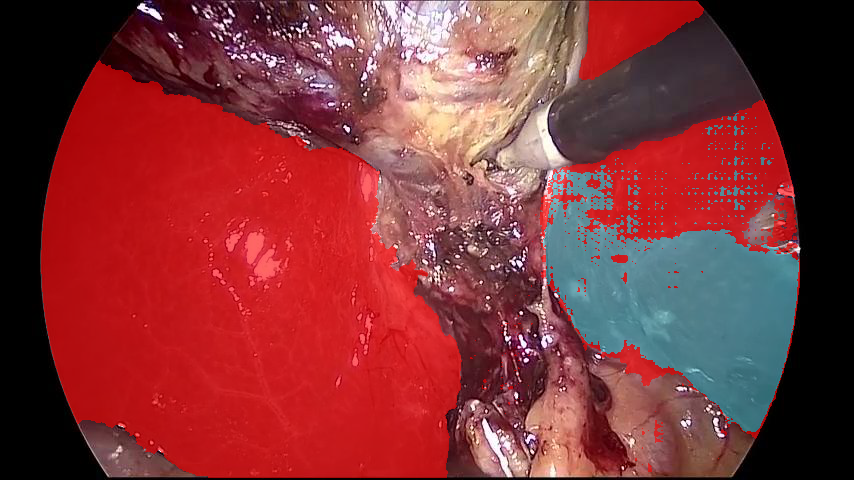}
\includegraphics[width=0.32\textwidth]{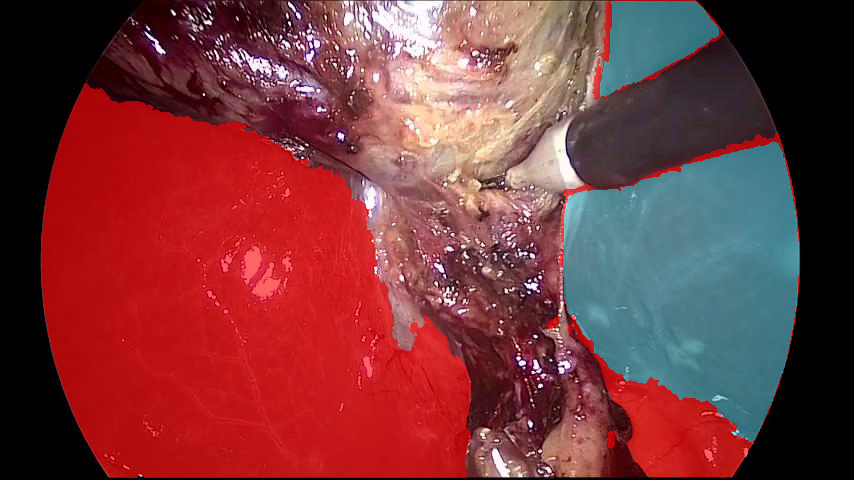}
\includegraphics[width=0.32\textwidth]{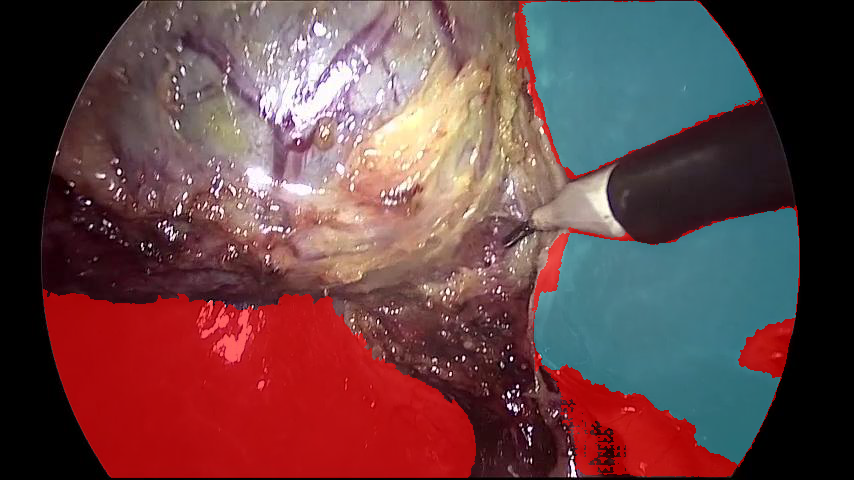}
\includegraphics[width=0.32\textwidth]{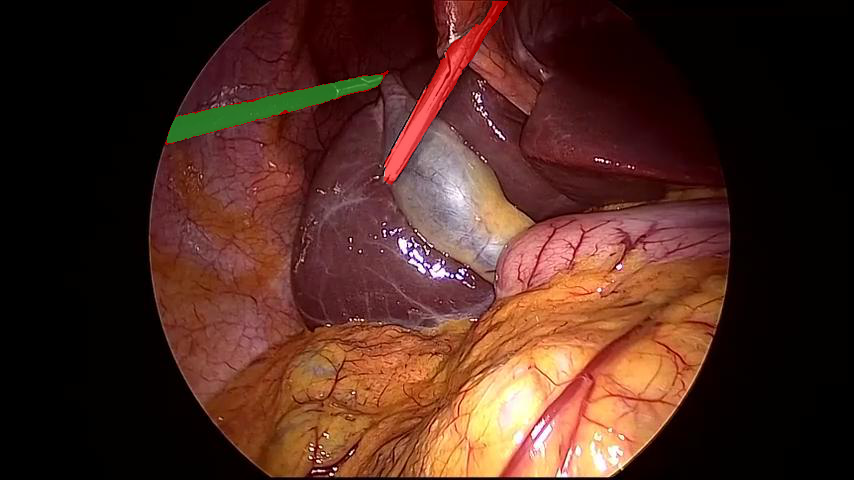}
\includegraphics[width=0.32\textwidth]{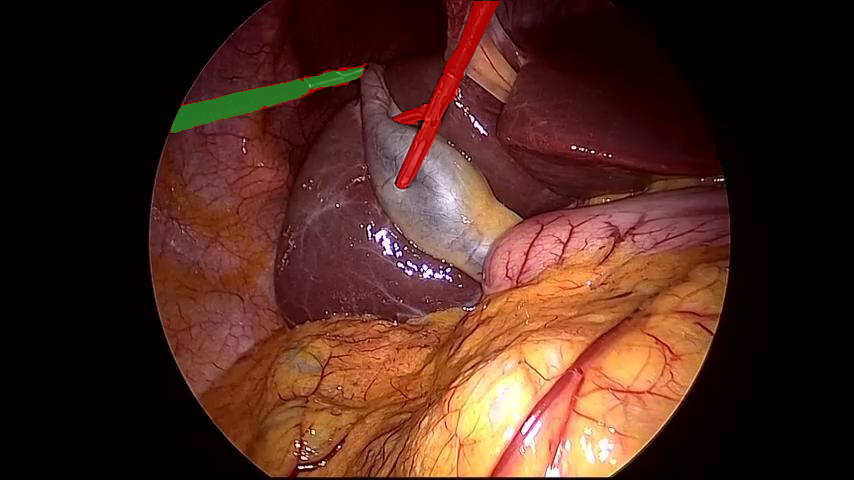}
\includegraphics[width=0.32\textwidth]{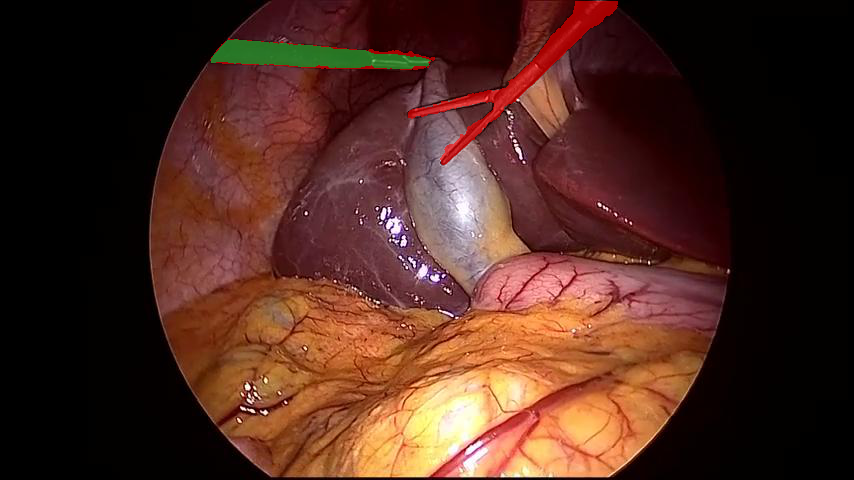}
\caption{Predictions on recipient patient videos}
\end{subfigure}
\caption{(a) Tracking masks obtained from donor patients and (b) segmentation predictions obtained from SAM2 using tracking masks for recipient patients based on the tracked objects. The first row corresponds to the \textbf{\gbla{gallbladder}}, the second row to the \textbf{\liver{liver}}, and the third row to the \textbf{\grasper{surgical grasper}}. Incorrect predictions (false negatives and false positives) are highlighted in \textbf{\red{red}}. Tracked objects in recipient patients obtain an average \textbf{IOU score $\mathbf{<0.5}$} for their videos, demonstrating limitations of the approach under certain circumstances.}
\label{fig:bad_examples}
\end{figure}

In Figure~\ref{fig:good_examples} and Figure~\ref{fig:bad_examples}, we present several intriguing qualitative examples showcasing cases where donor annotation results are on par with or even exceed the baseline, as well as cases where the results are suboptimal, respectively. We now discuss several of these interesting cases in more detail.

\textbf{Fine-grained object recognition}. The second row of Figure~\ref{fig:good_examples} indicates that, despite the tracking mask covering only a small area of the liver, recipient videos achieve nearly perfect segmentation. This demonstrates SAM2's ability to generalize from limited reference information and accurately propagate object identity across frames.

\textbf{Error correction}. As shown in the bottom row of Figure~\ref{fig:good_examples}, initial predictions incorrectly segment the gallbladder along with the grasper. However, this error is automatically corrected in later frames, ultimately achieving state-of-the-art average IoU for this video.

\textbf{Generalized multi-object tracking}. The last row of Figure~\ref{fig:good_examples} shows that, despite the tracking mask containing only one grasper, both graspers in the recipient video are correctly identified. This is particularly important, as using frames from the recipient video would require annotating each grasper separately, potentially causing tracking issues if objects re-enter the scene during surgery.

\textbf{Confusion between similar structures}. The example in the first row of Figure~\ref{fig:bad_examples} shows that the model confuses gallbladder with connective and fat tissue. This is the primary reason for low performance for gallbladder in Table~\ref{tbl:iou_table} for donor frames. Without a detailed tracking objective for gallbladder, SAM2 confuses gallbladder with similar looking structures surrounding it.

\textbf{Segmentation failure due to occlusion}. The second row of Figure~\ref{fig:bad_examples} illustrates a case where excessive blood occludes the liver, preventing detection. Such occlusions disrupt segmentation consistency by hindering feature extraction, highlighting a limitation of donor-based tracking when faced with drastic visual changes.

\textbf{Inconsistent multi-object tracking}. While donor frames often enable successful identification of multiple graspers, the second grasper is occasionally missed (see the last row in Figure~\ref{fig:bad_examples}). This inconsistency weakens temporal coherence in segmentation, leading to tracking errors that may impact the reliability of automated surgical analysis.

\section{Conclusions and Directions for Future Research}

In this work, we explored the feasibility of an unconventional approach: using segmentation annotations from other patients as tracking frames to eliminate the need for human intervention in real-time surgical video segmentation. Our experiments demonstrated that this method can achieve segmentation accuracy comparable to, and in some cases exceeding, object tracking using a patient's own frames. These findings suggest that carefully selected donor frames can serve as effective alternatives, improving segmentation consistency and broadening the applicability of automated surgical video analysis in real-world settings. Additionally, we discussed scenarios where this approach proves beneficial and identified its failure modes, paving the way for future research.

Our findings in Table~\ref{tbl:iou_table} and Figure~\ref{fig:donor_variance} also reveal substantial variability in the effectiveness of donor tracking frames, where some frames achieve state-of-the-art performance while others result in subpar segmentation. This variability highlights the importance of selecting optimal donor frames to maximize segmentation accuracy. As a result, our work paves the way for future research focused on donor frame selection strategies, including the development of automated methods to identify the most effective frames based on similarity metrics, anatomical features, or deep learning-based selection criteria. Improving donor frame selection could further enhance segmentation consistency and reduce performance variability, making this approach more reliable for real-time surgical video analysis.

\bibliographystyle{splncs04}
\bibliography{main}
\end{document}